\pdfoutput=1
\documentclass[11pt]{article}

\usepackage[final]{acl}

\usepackage{times}
\usepackage{latexsym}
\usepackage{multicol}
\usepackage{multirow}
\usepackage[T1]{fontenc}
\usepackage[utf8]{inputenc}

\usepackage{microtype}

\usepackage{inconsolata}

\usepackage{graphicx}
\usepackage{booktabs}
\usepackage{diagbox}
\usepackage{array} 
\usepackage{makecell}
\usepackage{amsmath}
\usepackage{amsfonts}
\usepackage{hyperref}
\usepackage{amssymb}
\usepackage{pifont}
\usepackage[ruled,linesnumbered]{algorithm2e}
\usepackage{xcolor}%

\usepackage{marvosym} 
\title{DICE: Structured Reasoning in LLMs through SLM-Guided Chain-of-Thought Correction}

\author{
 \textbf{Yiqi Li\textsuperscript{1}},
 \textbf{Yusheng Liao\textsuperscript{1,2}},
 \textbf{Zhe Chen\textsuperscript{1,2}},
 \textbf{Yanfeng Wang\textsuperscript{1,2}},
 \textbf{Yu Wang\textsuperscript{1,2}\thanks{Corresponding author}}
 \\
\\
 \textsuperscript{1}Shanghai Jiao Tong University,\\
 \textsuperscript{2}Shanghai Artificial Intelligence Laboratory
\\
 \small{
   \texttt{\{17-adamant,liao20160907,chenzhe2018,wangyanfeng,yuwangsjtu\}@sjtu.edu.cn}
 }
}

\begin{document}

\maketitle
\begin{abstract}
When performing reasoning tasks with user-specific requirements, such as strict output formats, large language models (LLMs) often prioritize reasoning over adherence to detailed instructions. Fine-tuning LLMs on supervised datasets to address this is impractical due to high computational costs and limited parameter access. To tackle this, we propose DICE, a lightweight framework that guides small language models (SLMs) to refine LLMs' outputs through chain-of-thought (CoT) correction. DICE decouples the process by first prompting LLMs to generate natural language responses, then using trained SLMs to analyze and refine these outputs to meet structured output specifications. This framework preserves LLMs' broad knowledge and reasoning capabilities while ensuring the outputs conform to user demands. 
Specifically, DICE first constructs structured CoT adaptation datasets via a two-stage method and subsequently applies a dual-tuning strategy to fine-tune SLMs for generating structured outputs in an analyze-then-answer pattern.\footnote{The datasets and code will be available at \href{https://github.com/1717Li/DICE}{https://github.com/1717Li/DICE}}
Experiments demonstrate that DICE improves the average format accuracy and content correctness of LLM outputs by 35.4\% and 29.4\%, respectively, achieving state-of-the-art (SOTA) performance over other competitive baselines.

\end{abstract}

\section{Introduction}

\begin{figure}[ht]
    \centering
    \includegraphics[width=\linewidth]{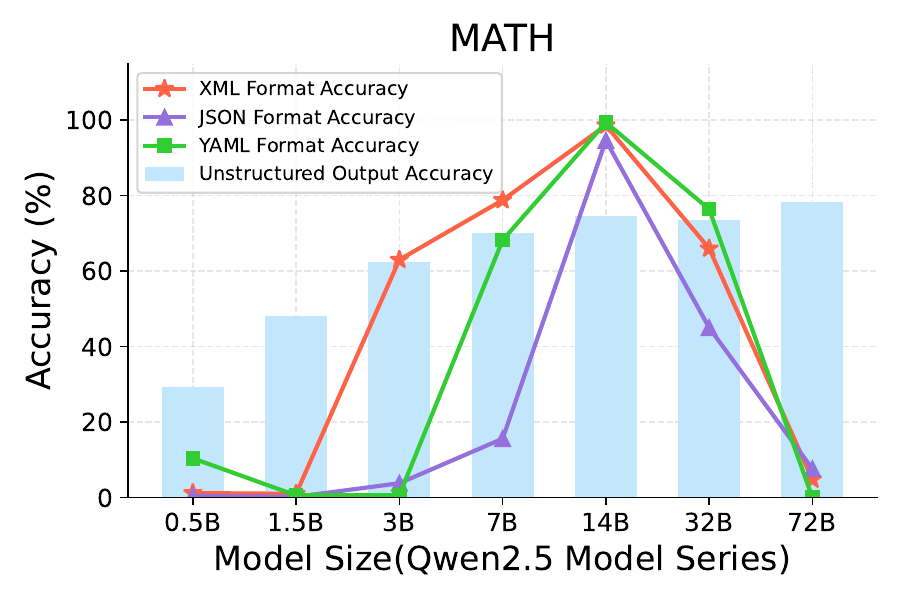}
    \caption{\textbf{Structured format accuracy and unstructured output accuracy across model sizes on \textit{MATH}.} The models are required to generate structured output given 2-shot prompts. The bars represent content accuracy of unstructured natural language outputs, and the lines denote the format accuracy of structured outputs. More details about formats are in Appendix \ref{sec:format}. 
    }
    \label{fig:math}
    
\end{figure}

\begin{figure*}[ht]
    \centering
    \includegraphics[width=\linewidth]{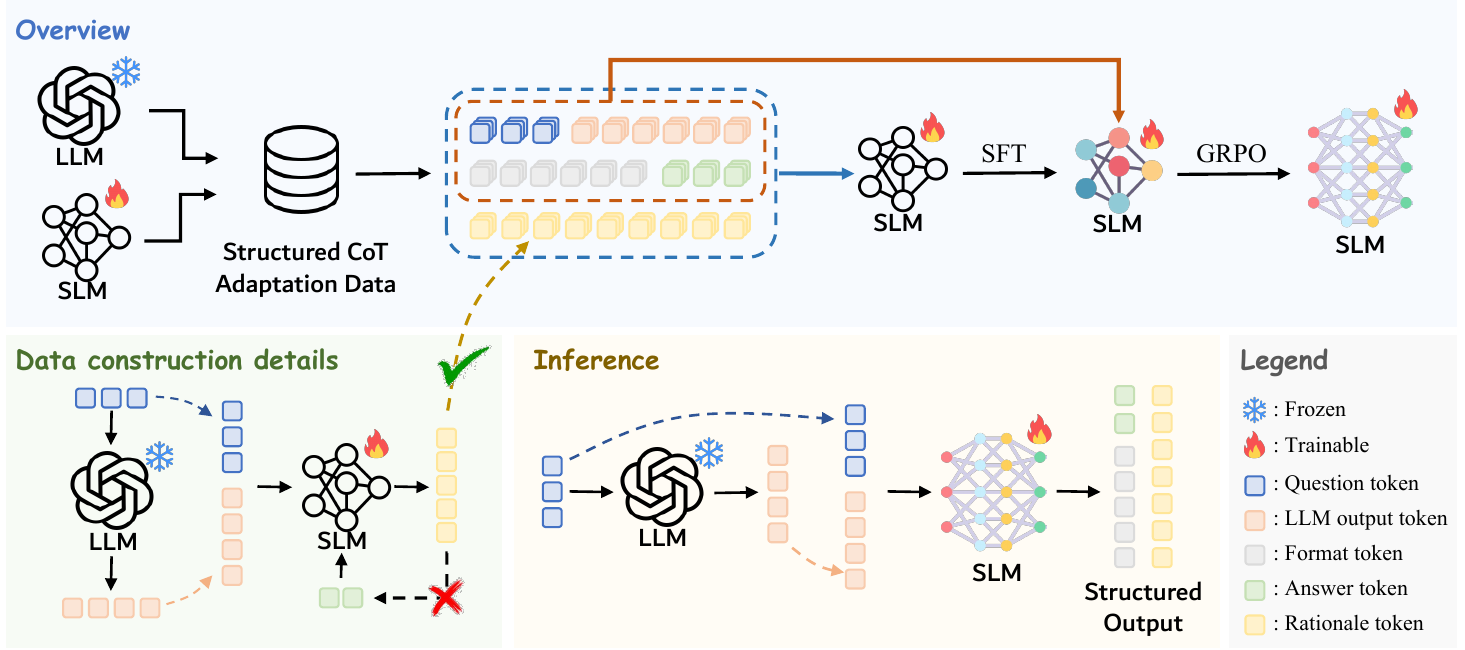}
    \caption{\textbf{Overview of DICE framework.} The training process comprises two sequential phases: DICE first employs a two-stage strategy to construct structured chain-of-thought data and subsequently implements a dual-tuning methodology to optimize the SLM to enforce rigorous format compliance. During inference, the trained SLM systematically analyzes and refines the natural language outputs from the LLM.}
    \label{fig:framework}
    
\end{figure*}
Large language models (LLMs) have demonstrated significant advancements across diverse natural language processing (NLP) tasks, exhibiting exceptional capabilities in language comprehension and reasoning \cite{guo2025deepseek,yang2024qwen2.5,grattafiori2024llama,team2024gemma,chen2025towards,hurst2024Gpt4o}. 
Their ability to follow general instructions is crucial in practical scenarios, such as complex decision-making, scientific research, and automated problem solving \cite{ouyang2022training,qin2024infobench,zhao2025incontext}. 
However, this instruction-following ability tends to degrade when LLMs are applied to challenging reasoning tasks \cite{tam2024let,shorten2024structuredrag,chenStructTestBenchmarkingLLMs2025}, which restricts the broader application in reasoning-intensive tasks.
% ~\ysl{Here needs some reference to explain why instruction-following capacity is important. Like `which is important in practical scenarios, such as complex decision-making, scientific research, and automated problem solving, where accurate reasoning guided by precise instructions is critical. This limitation restricts the broader application of LLMs in reasoning-intensive tasks, hindering their potential to fully assist in domains requiring both strong reasoning and reliable instruction adherence.'} 
% Therefore, it is important to improve the model's instruction-following ability while maintaining its reasoning ability.
% In this work, we focus on reasoning tasks with structured output requirement, such as XML and JSON, which are extensively utilized in real-world applications like agents and information extraction. As illustrated in Figure~ \ref{fig:math}, as the model's parameter size increases, its reasoning ability improves, resulting in an overall increase in the accuracy of the model's natural language outputs. 
% However, this does not imply that larger models inherently possess stronger instruction-following capabilities. For larger language models, such as 32B and 72B models, a greater portion of computational resources is allocated to reasoning on more challenging problems, which leads to less focus on the specific instructions provided by the user. As a result, the format accuracy of the model output decreases.

 While scaling up LLMs can enhance their reasoning capacity \cite{kaplan2020scalinglaw,zhong2021largerpretrainedlanguagemodels,naveed2024comprehensiveoverviewlargelanguage}, we observe a counterintuitive trade-off: larger models sometimes exhibit weaker adherence to user-specific instructions compared to smaller counterparts, even when their underlying reasoning is correct. 
% This phenomenon can be attributed to the interplay between computational resource allocation and the inherent complexity of multitasking during autoregressive decoding.~\ysl{Here needs citations, otherwise it can not use the word `attributed to'.}
% As model size increases, a greater proportion of its generative bandwidth is devoted to solving the core reasoning problem—such as mathematical derivations or logical inference—leaving fewer resources for syntactic precision in structured outputs.~\ysl{Here needs citations. Otherwise, it is more like a metaphor than a fact.} This aligns with broader observations that LLMs under cognitive load prioritize high-level task completion over low-level formatting. Larger models, optimized for broad task performance, may inadvertently deprioritize strict format compliance, treating it as secondary to semantic correctness. This suggests a task interference effect: as reasoning demands grow, the model’s ability to rigidly adhere to syntactic constraints diminishes. 
Our preliminary experiments (Figure~\ref{fig:math}) provide empirical evidence of this trade-off, focusing on user instructions related to specific output formats. It is observed that format accuracy peaks at mid-scale models while declining in larger models, despite their superior reasoning performance. 
Specifically, large-scale models (e.g., 32B and 72B parameters) optimized for diverse tasks tend to allocate more attention to solving difficult problems, but often at the expense of strict adherence to output formatting instructions.
The fragility of structured outputs exacerbates this issue—minor deviations (e.g., a misplaced bracket in JSON) can lead to complete parsing failures, disproportionately penalizing larger models during evaluation.

A natural solution is to fine-tune LLMs on the supervised dataset, but it is associated with several critical challenges: 
(1) \textbf{Inefficiency}: fine-tuning LLMs typically requires prohibitive computational resources and extended training duration; (2) \textbf{Alignment Tax}: task-specific fine-tuning risks catastrophic forgetting, which can inadvertently lead to performance degradation \cite{ouyang2022training,bubeck2023sparks,jiang2024taia}. (3) \textbf{Impracticality}: for many API-only LLMs such as GPT-4 \cite{achiam2023gpt}, fine-tuning is infeasible due to inaccessible of model parameters.
Recent studies have investigated collaborative frameworks that utilize small language models (SLMs) to effectively adapt LLMs to domain-specific tasks. Some methods use the probability distribution shift of SLM during fine-tuning to calibrate the LLM outputs \cite{liu2024proxy,ormazabal2023comblm}, while others employ model collaboration techniques to facilitate multi-step reasoning generation and path search \cite{sun2024bbox,fan2025gboost,kim2025guiding,zheng2025citercollaborativeinferenceefficient}. Additionally, several studies fine-tune SLMs to learn the correctional residuals between the ground-truth and the LLM-generated answers \cite{ji2024aligner,kim2024cobb,chen2024improvinglargemodelssmall}. 
However, these methods originally focus on enhancing the LLM's reasoning performance while overlooking its capability to follow instructions. 
Moreover, they fail to fully exploit the information embedded in the outputs of LLM, resulting in a high mis-correction rate. 
Thus, even when applied to structured reasoning tasks, these methods fail to adequately balance format output and reasoning performance.
% these approaches struggle to effectively integrate the reasoning capabilities of LLMs with the fine-tuning advantages of SLMs, often resulting in issues such as mis-correction.

To address these limitations, we propose a framework that adapts LLMs to structured reasoning tasks by guiDing SLMs to thInk with Chain-of-thought corrEction (DICE), as illustrated in Figure~\ref{fig:framework}. The framework operates in two stages: first, the LLM is prompted to produce unstructured natural language responses, avoiding interference from complex formatting requirements that could degrade reasoning quality. Then, a trained SLM is deployed to refine these outputs into specific formats. To train the SLM, we employ a two-stage process to generate rationales and construct structured chain-of-thought (CoT) adaptation datasets, followed by a dual-tuning strategy that guides the SLMs to perform deep analysis on LLM outputs before generating final answers. The core innovation of DICE lies in our novel use of the model collaboration framework to enhance the instruction-following capabilities of LLMs and the analyze-then-answer pattern used in SLMs generation. By utilizing chain-of-thought prompting to stimulate the reasoning ability of SLMs \cite{wei2023chainofthoughtpromptingelicitsreasoning,lyu2023faithfulchainofthoughtreasoning,srivastava2025reasoningabilitysmalllanguage}, they are able to improve instruction-following ability without compromising the inherent reasoning performance of LLMs, thereby addressing the mis-correction issues observed in prior approaches.

We conduct extensive experiments on five reasoning benchmarks to validate DICE’s effectiveness in adapting LLMs to downstream structured tasks. Compared to LLM with a 2-shot prompt, DICE achieves significant improvements, demonstrating average gains of $35.4\%$ in format accuracy and $29.4\%$ in content accuracy. Moreover, DICE consistently outperforms other baselines across nearly all evaluated datasets. Our key contributions can be summarized as follows:
\begin{itemize}
    \item To the best of our knowledge, we are the first to identify the negative correlation between instruction-following ability and model scale in reasoning tasks: while larger models exhibit stronger reasoning capabilities, their adherence to instructions tends to decline.
    \item We introduce DICE, a lightweight framework that leverages SLM to adapt LLMs to structured reasoning tasks. DICE operates without modifying the LLMs' parameters, thereby circumventing the “alignment tax” associated with fine-tuning and preserving the LLMs' general knowledge. 
    \item Extensive experiments show that DICE outperforms other baselines in improving instruction-following capabilities in reasoning tasks without compromising reasoning performance. Furthermore, DICE demonstrates superior generalizability across datasets and models, making it applicable in a wider range of scenarios.
\end{itemize}

\section{Related Work}
\subsection{Instruction-Following Ability of LLMs}
The capability of LLMs to follow user instructions is critical for practical applications. In recent years, numerous studies on model instruction-following have emerged. IFEval \cite{zhou2023IFEval} and CIF-Bench \cite{li2024cif-bench} contain various instructions for evaluating the general instruction-following proficiency. FOFO \cite{xia2024fofo} and StructuredRAG \cite{shorten2024structuredrag} specifically target format compliance evaluation. Specifically, for tasks requiring structured outputs, constrained decoding-based methods \cite{willard2023efficientguidedgenerationlarge,koo2024automatabasedconstraintslanguagemodel,dong2025xgrammarflexibleefficientstructured} have been proposed to enforce models to generate responses in specific formats. However, such methods suffer from poor flexibility and simultaneously degrade the quality of model-generated content \cite{tam2024let}. To solve these problems, we first propose reasoning tasks with specific output formats to simultaneously evaluate model instruction-following and reasoning abilities. Secondly, we introduce a model collaboration-based approach that enhances the model's instruction-following ability while improving its reasoning performance.

\subsection{Collaboration of SLMs and LLMs}
Due to prohibitively high computational costs and inaccessibility to model parameters, direct fine-tuning of LLMs remains infeasible for most researchers. This challenge has driven extensive exploration into LLM and SLM collaborative frameworks for task-specific adaptation. Distribution alignment approaches \cite{liu2024proxy,ormazabal2023comblm} attempt to integrate the output distribution shifts during SLM fine-tuning with LLM output distributions, but their practical applicability is circumscribed by the inaccessibility of full vocabulary probability distributions of various LLMs. Routing-based mechanisms \cite{sun2024bbox,aggarwal2024automix,fan2025gboost} decompose tasks into multi-step reasoning processes, where SLM selects optimal paths on multiple responses generated by LLM or dynamically selects between SLM and LLM during inference. However, these methods lead to increased computational cost and latency due to repeated LLM invocations. SLM correction frameworks \cite{kim2024cobb,ji2024aligner,kim2025guiding} train SLM to learn the residual between the LLM output and target answer, but existing methods suffer from high mis-correction rates due to insufficient analysis and utilization of LLM output. When applied to structured reasoning tasks, these challenges persistently undermine the model's ability to maintain structured output fidelity while ensuring reasoning accuracy in complex questions. To address these limitations, this work constructs structured chain-of-thought adaptation benchmarks and leverages the analyze-then-answer pattern to enhance the reasoning and correction capability of SLMs.

\section{Method}
In this section, we delve into the technical details of adapting LLMs to structured reasoning tasks by guiDing SLMs to thInk with Chain-of-thought corrEction (DICE). In Section~\ref{sec:data-cons}, we present the two-stage construction process of the structured chain-of-thought adaptation dataset. Next, in Section~\ref{sec:fine-tune}, we introduce the dual-tuning strategy used to fine-tune the SLM. The overview of DICE is presented in Figure~\ref{fig:framework} and Algorithm \ref{alg:AOS}.

\begin{algorithm}[ht]
    \small
    \caption{Algorithm of DICE}
    \label{alg:AOS}
    \KwIn{Pretrained LLM $\mathcal{M}$, SLM $\pi_{\theta}$, original dataset $\mathcal{D} = \{(x^i,y^i)\}_{i=1}^N$, format token $y_f$, learning rate for SFT $\eta_S$, learning rate for SFT $\eta_G$, training iteration for SFT $T_S$, training iteration for GRPO $T_G$}
    \vspace{2pt}\hrule\vspace{4pt} 
    % \normalem
    \tcp{Structured $\mathcal{Q}$ construction}
    $\mathcal{Q}\gets \emptyset$\;
    \For {$x^i,y^i \in \mathcal{D}$}{
        $y_{o}^{i} \sim \mathcal{M}(\cdot|x^i)$\\
        $r^{i}_1,y^{i}_{s,1} \sim \pi_0(\cdot|x^i,y_o^i)$\\
        \eIf {\text{$y^{i}_{s,1}=y^i$}}{
        $\mathcal{Q} \gets \mathcal{Q}\cup\{(x^i,y^i_o,y_f,r^i_1,y^i)\}$
        }
        {$r^{i}_2,y^{i}_{s,2} \sim \pi_0(\cdot|x^i,y_o^i,y^i)$\\
        \If {$y^{i}_{s,2}=y^i$}{
            $\mathcal{Q} \gets \mathcal{Q}\cup\{(x^i,y^i_o,y_f,r^i_2,y^i)\}$
            }
        }      
    }
    \tcp{Fine-tuning the SLM}
    \For {$t=1$ to $T_S$}{
    $\mathcal{L}_{SFT}(\theta,\mathcal{Q}) \gets \mathbb{E}_{\mathcal{Q}}[\mathcal{L}_{SFT}(\theta)]$ (Eq. \ref{eq:l1}) \\
    $\theta \gets \theta-\eta_S\nabla_{\theta} \mathcal{L}_{SFT}(\theta,\mathcal{Q}) $
    }
    \For {$t=1$ to $T_G$}{
    $\mathcal{L}_{GRPO}(\theta,\mathcal{Q}) \gets \mathbb{E}_{\mathcal{Q}}[\mathcal{L}_{GRPO}(\theta)]$ (Eq. \ref{eq:grpo}) \\
    $\theta \gets \theta-\eta_S\nabla_{\theta} \mathcal{L}_{GRPO}(\theta,\mathcal{Q}) $
    }
    \KwOut{Fine-tuned SLM $\pi_{\theta}$}
\end{algorithm}

\subsection{Structured Chain-of-Thought Adaptation Dataset Construction} \label{sec:data-cons}
%two-stage construction process, formulas

Given a pre-trained LLM $\mathcal{M}$, pre-trained SLM $\pi_0$, and question-answer pairs $\mathcal{D} = \{(x^i,y^i)\}_{i=1}^N$ from the original training set ($N$ is the size of the benchmark), our approach begins by instructing the LLM $\mathcal{M}$ directly to obtain the natural language outputs:
\begin{equation}
    y^i_o \sim \mathcal{M}(\cdot|x^i)
\end{equation}

To address the problem of high mis-correction rate in prior approaches, we construct analytical data based on $y_o$, which can guide the SLM to reason before generating the final answers. To reduce computational cost and generate rationales more suitable for SLM to learn, we propose a two-stage methodology, akin to STaR \cite{zelikman2022star}, to instruct the pre-trained SLM $\pi_0$ to reason the LLM output and provide the predicted answer. Considering the limited instruction-following capability of $\pi_0$, we first generate two demonstrations with the assistance of LLM in the following two steps. In the first stage, the rationale $r_1^i$ and predicted answer $y^i_{s,1}$ are formulated as:
\begin{equation}
    (r_1^i,y^i_{s,1})\sim \pi_0(\cdot|x^i,y_o^i)
\end{equation}

Based on the assumption that rationales leading to correct predicted answers possess positive utility, we filter the generated outputs, retaining only those associated with accurate answers ($y^i_{s,1}=y^i$). 
In the subsequent stage, for the filtered-out samples, where $\pi_0$ alone struggles to generate meaningful rationales, we append the answer label $y^i$ as a contextual hint to the original input of these challenging samples, then regenerate rationales and answers:
\begin{equation}
    (r_2^j,y^j_{s,2})\sim \pi_0(\cdot|x^j,y_o^j,y^j), \text{for}\ y^j_{s,1}\neq y^j
\end{equation}

After generation, we repeat the aforementioned filtering procedure. Empirical results from our experiments demonstrate that, following this two-stage approach, the SLM successfully generates rationales yielding correct predictions for over $90\%$ of the original training split. Consequently, we discard the filter-out samples in the second stage.
The final rationale set can be formulated as $\mathcal{R}=\mathcal{R}_1\cup\mathcal{R}_2=\{r_1^i|y^i_{s,1}=y^i\}_{i=1}^{N_1}\cup\{r_2^j|y^j_{s,2}=y^j\}_{j=2}^{N_2}$, where $N_1$ and $N_2$ denote the number of samples retaining after each filtering procedure. Ultimately, we embed the rationales and answers into the required format (presented as $y_f$), obtaining the new training target $t^k=(y_f,r^k,y^k)$. The final structured chain-of-thought adaptation dataset is $\mathcal{Q}=\{(x^k,y_o^k,t^k)\}_{k=1}^{N_1+N_2}$. 
% With this dataset, the SLM can be trained to conduct a thorough analysis of the LLM's responses before generating the answer, thereby activating the SLM’s reasoning ability and effectively leveraging the output from the LLM.

\subsection{Guiding SLMs to Think Through a Dual-Tuning Strategy}\label{sec:fine-tune}
The most straightforward fine-tuning approach is SFT, wherein the $\pi_0$ is trained to predict all target tokens with equal emphasis. The training objective is to minimize the cross-entropy loss:
\begin{equation}\label{fun:sft-loss}
    \mathcal{L}_{SFT}(\theta) = -\log \pi_{\theta}(t|x,y_o) 
\end{equation}
where $(x,y_o,t)\sim\mathcal{Q}$. The target $t$ consists of three components, allowing us to decompose the loss function and gradient into three corresponding terms:
\begin{equation}
\begin{aligned}\label{eq:l1}
    \mathcal{L}_{SFT}(\theta) &= -\log \pi_{\theta}(y_f,r,y|x,y_o) \\
    &= \mathcal{L}_f(\theta) + \mathcal{L}_r(\theta) + \mathcal{L}_y(\theta)
\end{aligned}
\end{equation}
\begin{equation}
    \resizebox{.88\hsize}{!}{$\nabla_{\theta} \mathcal{L}_{SFT}(\theta)=\nabla_{\theta} \mathcal{L}_f(\theta) + \nabla_{\theta}\mathcal{L}_r(\theta)+ \nabla_{\theta} \mathcal{L}_y(\theta)$}
\end{equation}

In practice, these three components are intrinsically interwoven and vary in length across actual outputs, rendering separate computation infeasible. Moreover, the rationale $r$ typically constitutes the longest segment; for example, in the MATH dataset requiring XML output, the average token ratio between $y_f$, $r$, and $y$ is approximately $25:135:1$. Consequently, during gradient computation, $\lvert\nabla_{\theta} \mathcal{L}_r(\theta)\rvert$ can significantly exceed $\lvert\nabla_{\theta} \mathcal{L}_f(\theta)\rvert$ and $\lvert\nabla_{\theta} \mathcal{L}_y(\theta)\rvert$. This imbalance causes $\pi_0$ to prioritize minimizing $\mathcal{L}_r(\theta)$ during optimization, resulting in insufficient learning of $y_f$ and $y$. Meanwhile, since $r$ is generated by $\pi_0$ itself, it primarily contains in-domain knowledge that provides limited additional information. In contrast, novel knowledge such as user instruction and task-specific information is primarily contained in $y_f$ and $y$, which necessitate more focused learning. 

To address this challenge, we propose a dual-tuning strategy to progressively optimize the SLM. First, we conduct SFT utilizing Low-Rank Adaptation (LoRA \cite{hu2022lora}) on $\pi_0$ to rapidly acquire format specifications and the analyze-then-answer generation pattern, obtaining $\pi_{SFT}$. Subsequently, we employ a more granular fine-tuning method to further optimize $\pi_{SFT}$ using the GRPO \cite{shao2024grpo} algorithm (detailed in Appendix \ref{ap:algorithm}). For model output $\hat{t}=(\hat{y_f},\hat{r},\hat{y})$, we design reward functions that assign rewards solely based on $\hat{y_f}$ and $\hat{y}$, neglecting $\hat{r}$. The total reward is calculated as follows:
\begin{equation}
\text{reward}= \begin{cases}
2, &\text{both $\hat{y_f}$ and $\hat{y}$ are correct} \\
1, &\text{one of $\hat{y_f}$ and $\hat{y}$ is correct} \\
0, &\text{both $\hat{y_f}$ and $\hat{y}$ are incorrect}
\end{cases} 
\end{equation}
\begin{table*}[ht]
\centering
\resizebox{1\linewidth}{!}{
\begin{tabular}{cclcccccccccccc}
\toprule

\multirow{2}{*}{\textbf{SLM}}&\multirow{2}{*}{\textbf{LLM}} & \multirow{2}{*}{\textbf{Method}} & \multicolumn{2}{c}{\textbf{GSM8K}} & \multicolumn{2}{c}{\textbf{MATH}} & \multicolumn{2}{c}{\textbf{CSQA}} & \multicolumn{2}{c}{\textbf{MedQA-zh}} & \multicolumn{2}{c}{\textbf{StrategyQA}} & \multicolumn{2}{c}{\textbf{\textbf{Average}}}\\ 
\cmidrule(rl){4-5} \cmidrule(rl){6-7} \cmidrule(rl){8-9} \cmidrule(rl){10-11} \cmidrule(rl){12-13} \cmidrule(rl){14-15}
              &  & &  F-Acc & C-Acc & F-Acc & C-Acc & F-Acc & C-Acc & F-Acc & C-Acc & F-Acc & C-Acc & F-Acc & C-Acc \\ 
                % & & (\%) & (\%) &(\%) &(\%) &(\%) &(\%) &(\%) &(\%) &(\%) &(\%) &(\%) &(\%) \\
                \cmidrule{1-15}
\multirow{3}{*}{\ding{55}} 
        & 72B & 0-shot & 0.0 & 0.0 & 0.2 & 0.0 & 0.0 & 0.0 & 0.0 & 0.0 & 0.0 & 0.0 & 0.0 & 0.0 \\
        & 72B & ICL & 64.2 & 61.8 & 4.8 & 4.2 & 97.4 & 82.6 & 77.0 & 67.2 & 95.2 & 74.2 & 67.7 & 58.0 \\
        & 72B & Reflection & 64.4 & 62.0 & 4.8 & 4.2 & 97.4 & 82.6 & 77.2 & 67.4 & 95.2 & 74.2 & 67.8 & 58.1 \\
\cmidrule(rl){1-15}
\multirow{7}{*}{0.5B} 
        & \ding{55} & 0-shot &0.0 & 0.0 & 0.0 & 0.0 & 0.0 & 0.0 & 8.8 & 0.2 & 0.0 & 0.0 & 1.8 & 0.0 \\
        & \ding{55} & ICL & 10.8 & 2.6 & 1.2 & 0.4 & 76.2 & 14.4 & 57.8 & 8.2 & 84.3 & 47.6 & 46.1 & 14.6 \\
        & \ding{55} & SFT & \underline{98.6} & 25.6 & 94.6 & 12.4 & \textbf{100.0} & 57.6 & \textbf{100.0} & 45.2 & 96.9 & 59.0 & 98.0 & 40.0 \\
        &72B & Aligner & \textbf{99.6} & 46.2 & 96.2 & 48.6 & \textbf{100.0} & 78.6 & \textbf{100.0} & 86.8 & 96.9 & 69.4 & \underline{98.5} & 65.9 \\
        &72B& BBox-Adapter & 91.0 & 83.8 & 11.2 & 9.4 & \underline{99.4} & \underline{85.2} & 94.8 & 83.8 & 95.2 & \textbf{79.0} & 78.3 & 68.2 \\
        &72B& CoBB & 98.4 & \underline{94.2} & \underline{96.6} & \underline{76.6} & 97.4 & 79.6 & \underline{99.8} & \underline{84.0} & \underline{99.6} & 71.2 & 98.4 & \underline{81.1} \\
        &72B& DICE (Ours) & \textbf{99.6} & \textbf{95.2} & \textbf{99.4} & \textbf{79.0} & \textbf{100.0} & \textbf{85.8} & \textbf{100.0} & \textbf{88.0} & \textbf{100.0} & \underline{78.2} & \textbf{99.8} & \textbf{85.2} \\   
\cmidrule(rl){1-15}
\multirow{7}{*}{1.5B} 
        & \ding{55}& 0-shot & 0.0 & 0.0 & 0.4 & 0.2 & 7.4 & 0.0 & 13.8 & 0.0 & 0.0 & 0.0 & 4.3 & 0.0 \\
        & \ding{55}& ICL & 77.4 & 46.0 & 1.0 & 0.6 & 81.2 & 31.0 & \underline{99.0} & 31.4 & \underline{98.3} & 53.7 & 71.4 & 32.5 \\
        & \ding{55}& SFT & \underline{98.8} & 46.4 & 98.0 & 25.8 & \textbf{100.0} & 70.8 & \textbf{100.0} & 71.2 & \underline{98.3 }& 65.9 & 99.0 & 56.0 \\
        &72B& Aligner &  98.6 & 51.0 & \underline{98.2} & 38.6 & \textbf{100.0} & 79.2 & \textbf{100.0} & 80.6 & \textbf{100.0} & 73.8 & \underline{99.4} & 64.6 \\
        &72B& BBox-Adapter & 93.4 & 87.0 & 11.4 & 9.6 & \underline{99.0} & \underline{85.0} & 98.4 & \textbf{88.2} & 94.8 & \underline{78.6} & 79.4 & 69.7 \\
        &72B& CoBB & 97.4 & \underline{92.8} & 96.0 & \underline{74.8} & 98.6 & 82.4 & \underline{99.0} & 83.6 & \textbf{100.0} & 74.7 & 98.2 & \underline{81.7} \\
        &72B& DICE (Ours) & \textbf{99.8} & \textbf{95.6} & \textbf{99.6} & \textbf{79.8} & \textbf{100.0} & \textbf{85.6} & \textbf{100.0} & \underline{87.8} & \textbf{100.0} & \textbf{79.5} & \textbf{99.9} & \textbf{85.7} \\
\cmidrule(rl){1-15}    
\multirow{7}{*}{3B} 
        & \ding{55}& 0-shot & 82.0 & 63.8 & 60.6 & 38.6 & 89.6 & 1.4 & 54.2 & 0.0 & 84.3 & 55.9 & 74.1 & 31.9 \\
        & \ding{55}& ICL & 92.6 & 75.0 & 63.0 & 37.0 & 96.6 & 65.0 & 94.2 & 52.0 & 98.3 & 62.0 & 88.9 & 58.2 \\
        & \ding{55}& SFT & 99.4 & 63.4 & \underline{99.0} & 32.0 & \textbf{100.0} & 77.6 & \textbf{100.0} & 73.6 & \textbf{100.0} & 69.9 & \underline{99.7} & 63.3 \\
        &72B& Aligner & \textbf{99.8} & 61.8 & 98.8 & 50.6 & \textbf{100.0} & 80.0 & \textbf{100.0} & 83.6 & 99.1 & 74.2 & 99.5 & 70.0 \\
        &72B& BBox-Adapter & 93.2 & 85.8 & 10.6 & 8.8 & \underline{99.8} & \textbf{85.6} & 98.8 & \textbf{88.6} & 95.2 & \underline{78.6} & 79.5 & 69.5 \\
        &72B& CoBB & 97.0 & \underline{92.2} & 96.4 & \underline{74.6} & 98.4 & 83.4 & \underline{99.6} & 86.4 & \underline{99.6} & 73.4 & 98.2 & \underline{82.0} \\
        &72B& DICE (Ours) & \underline{99.6} & \textbf{94.2} & \textbf{99.6} & \textbf{77.8} & \textbf{100.0} & \underline{84.4} & \textbf{100.0} & \underline{88.0} & \underline{99.6} & \textbf{80.4} & \textbf{99.8} & \textbf{85.0} \\
        
\bottomrule
\end{tabular}
}
\caption{\textbf{Performance comparison on five downstream reasoning tasks with XML output requirement.} All models originate from the Qwen2.5 model series. The base LLM is Qwen2.5-72B-Instruct-GPTQ-Int4. For each model size of SLM, the highest and second-highest scores are highlighted in \textbf{bold} and \underline{underlined}, respectively.}
\label{tab:main}
\vspace{-1em}
\end{table*}

\section{Experiments}

\subsection{Experimental Setup}
\paragraph{Datasets and Metrics} We evaluate the effectiveness of DICE from four dimensions: mathematical reasoning (GSM8K \cite{cobbe2021gsm8k}, MATH \cite{hendrycks2021math}), commonsense reasoning (CommonsenseQA \cite{talmor2018commonsenseqa}, CSQA for short), domain-specific reasoning (MedQA-zh \cite{jin2021medqa}), and implicit reasoning (StrategyQA \cite{geva2021strategyqa}). To comprehensively evaluate the model's reasoning and instruction-following ability, we restructure the output into a more sophisticated XML format. We employ two metrics to evaluate the quality of the outputs: \textit{Format Accuracy} (F-Acc) and \textit{Content Accuracy} (C-Acc). Format Accuracy assesses adherence to structural elements and keywords prescribed by the template. Content accuracy is derived from the Exact Match (EM) score calculated for the final answer extracted from outputs. Notably, format compliance is a necessary condition for content accuracy. More details are in Appendix \ref{sec:data&metric}. 

\paragraph{Baselines} We compare DICE against three methodological categories: (1) \textit{Training-free method}: This category utilizes pre-trained large and small language models for response generation through 0-shot prompt and In-Context Learning (ICL). We also leverage the reflection baseline that feeds format-violated outputs from LLM ICL to the LLM and instruct it to reflect and regenerate the answer. (2) \textit{Supervised Fine-Tuning (SFT) method}: In this category, SLMs are directly fine-tuned on the supervised training data. (3) \textit{Model collaboration method}: Aligner \cite{ji2024aligner}, BBox-Adapter \cite{sun2024bbox}, and CoBB \cite{kim2024cobb}. Aligner leverages SLM to learn the mapping between LLM output and the ground-truth answer. BBox-Adapter scores iterative LLM generations via a trained evaluator, then applies beam search for optimal reasoning path selection. CoBB first constructs contrastive examples, then deploys SLM to learn from the pair-wise preference data through the ORPO \cite{hong2024orpo} algorithm. \textbf{It should be noted that we apply all methods to the structured reasoning tasks, even though their original works primarily focus on content refinement.}

\paragraph{Implementation Details} We select Qwen2.5-72B-Instruct-GPTQ-Int4 \cite{yang2024qwen2.5} as the LLM. The initial SLMs are derived from the instruction-tuned models within the Qwen2.5 series. For ICL, models generate responses through a 2-shot prompt. For the BBox-Adapter, we utilize a single generation step and classification mode. For CoBB, we generate one positive and one negative reasoning for each question. For both SFT and Aligner, the SLMs are trained for 3 epochs. For our proposed DICE, we initially fine-tune the SLMs using SFT for 2 epochs, followed by 1 epoch of GRPO fine-tuning. Further details on the implementation can be found in Appendix \ref{sec:implementation}.

\begin{table}[t]
\centering
\renewcommand\arraystretch{1.1}
\resizebox{0.95\linewidth}{!}{
\begin{tabular}{lcccc}
\toprule
\multirow{2}{*}{\textbf{Method}} & \multicolumn{2}{c}{\textbf{MATH JSON}} & \multicolumn{2}{c}{\textbf{MATH YAML}}\\ 
\cmidrule(rl){2-3} \cmidrule(rl){4-5}
    &  F-Acc &  C-Acc & F-Acc &  C-Acc  \\ \cmidrule{1-5}
    LLM (ICL) & 10.4 & 7.6 & 0.0 & 0.0 \\
    SLM (SFT) & 98.0 & 23.8 & 98.8 & 29.0 \\
    LLM + Aligner & 97.2 & 46.6 & 98.6 & 44.6 \\
    LLM + BBox-Adapter & 18.4 & 12.4 & 3.2 & 2.2 \\
    LLM + CoBB & 99.0 & 76.2 & 99.6 & 76.0 \\
    LLM + DICE(Ours) & \textbf{99.8} & \textbf{80.0} & \textbf{100.0} & \textbf{79.6} \\

\bottomrule
\end{tabular}
}
\caption{\textbf{Performance comparison on MATH specifying JSON and YAML output format.} The model size of the LLM and SLM utilized is 72B and 1.5B.}
\label{tab:dif-formats}
\vspace{-1em}
\end{table}

\subsection{Main Result}
Table~\ref{tab:main} presents the performance of DICE and other baselines on the five selected reasoning tasks with specialized XML output requirements. 
First, we observe that without any demonstrations, the Qwen2.5-72B-Instruct-GPTQ-Int4 largely fails to generate responses adhering to the specific formatting requirements specified in user instructions. When employing ICL, it exhibits substantial instability in format accuracy across diverse datasets. For instance, on more challenging benchmarks such as MATH, it tends to allocate excessive attention to problem-solving processes, consequently neglecting the formatting constraints outlined in instructions, results in substantially diminished formatting accuracy (below 5$\%$). Furthermore, in the reflection baseline, even after feeding the incorrect responses along with feedback back into the LLM for regeneration, the format accuracy shows little improvement. This indicates that relying solely on LLMs cannot simultaneously balance output format and reasoning performance for these questions.
However, after utilizing our proposed DICE framework with model size less than 3 billion parameters, near-perfect format accuracy (approaching $100\%$) is achieved consistently across all evaluated datasets, while simultaneously improving the content accuracy by an average of $29.4\%$ compared to LLM using ICL.
Moreover, under identical SLM size, DICE achieves either the best or second-best performance across all datasets, and obtains the highest average scores in both F-Acc and C-Acc. It significantly outperforms the collaboration-based baselines, including Aligner, BBox-Adapter, and CoBB, with average content accuracy gains of $18.4\%$, $16.1\%$, and $3.6\%$, respectively. 

For more structures such as JSON and YAML, we also conduct experiments with 1.5B SLM on the MATH dataset, with results summarized in Table \ref{tab:dif-formats}. It is observed that our proposed DICE consistently achieves the highest F-Acc and C-Acc for both JSON and YAML formats, outperforming other baseline approaches.
These results indicate that our proposed DICE framework not only adheres to user-specified output format constraints but also effectively harnesses the respective reasoning capabilities of both large and small language models, leading to high content accuracy.
\begin{table*}[ht]
\centering
\renewcommand\arraystretch{1.1}
\resizebox{1\linewidth}{!}{
\begin{tabular}{lcccccccccccc}
\toprule
\multirow{2}{*}{\textbf{Method}} & \multicolumn{2}{c}{\textbf{GSM8K}} & \multicolumn{2}{c}{\textbf{MATH}} & \multicolumn{2}{c}{\textbf{CSQA}} & \multicolumn{2}{c}{\textbf{MedQA-zh}} & \multicolumn{2}{c}{\textbf{StrategyQA}} & \multicolumn{2}{c}{\textbf{\textbf{Average}}}\\ 
\cmidrule(rl){2-3} \cmidrule(rl){4-5} \cmidrule(rl){6-7} \cmidrule(rl){8-9} \cmidrule(rl){10-11} \cmidrule(rl){12-13}
    & F-Acc & C-Acc & F-Acc & C-Acc & F-Acc & C-Acc & F-Acc & C-Acc & F-Acc & C-Acc  & F-Acc & C-Acc \\ 
    % & (\%) & (\%) &(\%) &(\%) &(\%) &(\%) &(\%) &(\%) &(\%) &(\%) &(\%) &(\%) \\
    \cmidrule{1-13}
    Qwen2.5-7B-Instruct (ICL) & 96.2 & 85.0 & 78.8 & 52.6 & 55.8 & 41.4 & 93.8 & 72.0 & 83.8 & 61.6 & 81.7 & 62.5 \\
    Qwen2.5-7B-Instruct + Aligner & 99.4 & 54.8 & 97.2 & 40.8 & \textbf{100.0} & 77.2 & \textbf{100.0} & 76.6 & \textbf{99.6} & 66.7 & 99.2 & 63.2 \\
    Qwen2.5-7B-Instruct + BBox-Adapter & 95.4 & 84.0 & 73.8 & 43.8 & 49.4 & 37.4 & 91.6 & 72.2 & 83.0 & 60.7 & 78.6 & 59.6 \\
    Qwen2.5-7B-Instruct + CoBB & 98.0 & 91.0 & 95.4 & 66.6 & 99.2 & 78.0 & 98.4 & 73.0 & 97.8 & 69.0 & 97.8 & 75.5 \\
    Qwen2.5-7B-Instruct + DICE & \textbf{99.8} & \textbf{92.2} & \textbf{99.4} & \textbf{71.0} & \textbf{100.0} & \textbf{78.6} & \textbf{100.0} & \textbf{81.4} & \textbf{99.6} & \textbf{70.7} & \textbf{99.8} & \textbf{78.8} \\
\cmidrule(rl){1-13}
    Meta-Llama-3-8B-Instruct (ICL) & 98.8 & 76.4 & 82.4 & 25.8 & 97.4 & 63.6 & 97.0 & 39.8 & 96.5 & 66.4 & 94.4 & 54.4 \\
    Meta-Llama-3-8B-Instruct + Aligner & 99.6 & 52.0 & 98.0 & 27.2 & \textbf{100.0} & \textbf{68.4} & \textbf{100.0} & \textbf{65.0} & \textbf{100.0} & 64.2 & \textbf{99.5} & 55.4 \\
    Meta-Llama-3-8B-Instruct + BBox-Adapter & 99.0 & 75.4 & 80.4 & 20.4 & 86.8 & 54.8 & 80.2 & 36.8 & 92.1 & 68.6 & 87.7 & 51.2 \\
    Meta-Llama-3-8B-Instruct + CoBB & 98.6 & 80.6 & 91.2 & 31.4 & 96.4 & 64.4 & 98.2 & 46.0 & 98.3 & 63.8 & 96.5 & 57.2 \\
    Meta-Llama-3-8B-Instruct + DICE & \textbf{100.0} & \textbf{81.6} & \textbf{99.6} & \textbf{33.2} & \textbf{100.0} & \textbf{68.4} & \textbf{100.0} & 48.6 & 97.8 & \textbf{69.0} & \textbf{99.5} & \textbf{60.2} \\
\cmidrule(rl){1-13}
    GPT-4.1-mini (ICL) & \textbf{100.0} & 96.2 & 78.6 & 67.4 & \textbf{100.0} & \textbf{82.8} & 97.8 & 76.2 & \textbf{100.0} & \textbf{83.0} & 95.3 & 81.1 \\
    GPT-4.1-mini + Aligner & 99.8 & 55.2 & 96.2 & 35.2 & \textbf{100.0} & 81.4 & \textbf{100.0} & 77.0 & 98.3 & 75.1 & 98.9 & 64.8 \\
    GPT-4.1-mini + BBox-Adapter & \textbf{100.0} & 92.0 & 81.4 & 65.6 & 98.2 & 81.0 & 96.6 & 74.0 & 99.6 & 82.1 & 95.2 & 78.9 \\
    GPT-4.1-mini + CoBB & 96.2 & 90.4 & 81.8 & 59.0 & 97.6 & 80.4 & 98.6 & 72.8 & 93.7 & 73.8 & 93.6 & 75.3 \\
    GPT-4.1-mini + DICE & \textbf{100.0} & \textbf{97.0} & \textbf{99.4} & \textbf{69.4} & \textbf{100.0} & \textbf{82.8} & \textbf{100.0} & \textbf{78.2} & \textbf{100.0} & 81.2 & \textbf{99.9} & \textbf{81.7} \\

\bottomrule
\end{tabular}
}
\caption{\textbf{Cross-model generalization ability of different methods.} The 1.5B SLMs of all methods are trained to adapt Qwen2.5-72B-Instruct-GPTQ-Int4 to reasoning tasks with XML format requirements. }
\label{tab:plug}
\vspace{-0.8em}
\end{table*}
\subsection{Generalizability Analysis}
\paragraph{Cross-Model Generalizability} Our DICE framework solely requires the outputs from the LLMs, making it applicable to diverse LLMs in a plug-and-play manner. To systematically evaluate the cross-model generalizability, we employ the 1.5B SLM (trained to adapt Qwen2.5-72B-Instruct-GPTQ-Int4 in Table \ref{tab:main}) to adapt other distinct LLMs: Qwen2.5-7B-Instruct \cite{yang2024qwen2.5}, Meta-Llama-3-8B-Instruct \cite{grattafiori2024llama}, and GPT-4.1-mini \cite{openai2025gpt41} to tasks with XML format requirements. As presented in Table \ref{tab:plug}, the results demonstrate that DICE can successfully adapt various large models to all XML-constrained reasoning tasks, achieving an average F-Acc exceeding $99.5\%$. Notably, DICE presents significant improvements in C-Acc compared to ICL baselines, with average gains of $16.3\%$ (Qwen2.5-7B-Instruct), $5.8\%$ (Meta-Llama-3-8B-Instruct), and $0.6\%$ (GPT-4.1-min). Furthermore, compared to other baselines, DICE achieves superior performance in both F-Acc and C-Acc, underscoring its strong capability for cross-model generalization.
\begin{figure}[t]
    \centering
    \includegraphics[width=\linewidth]{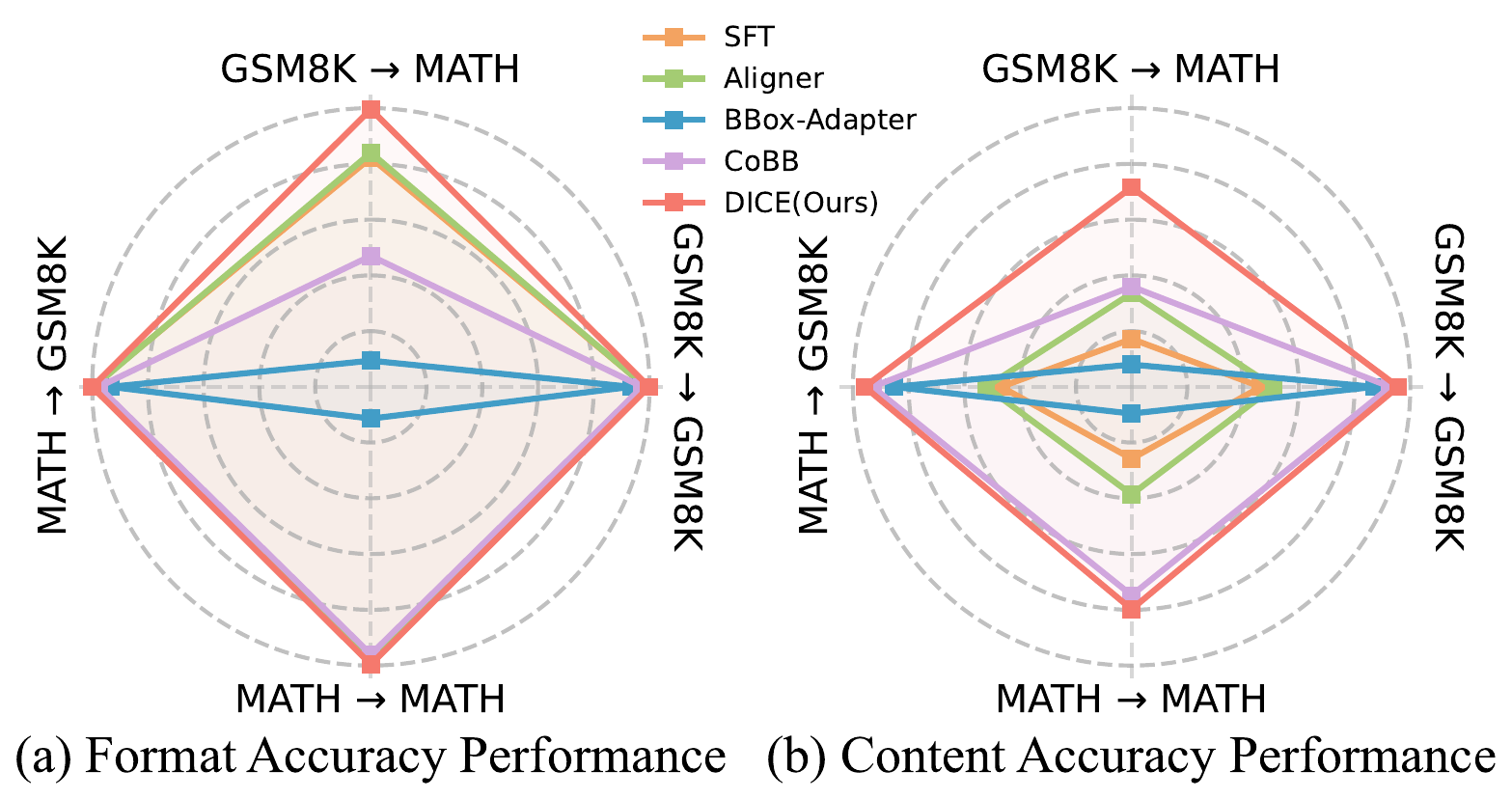}
    \caption{\textbf{Cross-dataset generalization ability of different methods.} The 1.5B SLMs trained on GSM8K and MATH through different methods are evaluated on test sets of both benchmarks. ``A$\rightarrow$B'' represents models that are trained on A and tested on B.}
    \label{fig:crossdata}
    \vspace{-1em}
\end{figure}
\paragraph{Cross-Dataset Generalizability}To further investigate the generalization performance of our method across datasets, we conduct cross-dataset validation experiments using the 1.5B SLM in Table \ref{tab:main} on GSM8K and MATH datasets. Specifically, apart from the consistent train and test datasets, we assess both cross-dataset generalization scenarios: (1) evaluating MATH test performance of models trained on GSM8K, and (2) evaluating GSM8K test performance of models trained on MATH. The experimental results are visualized in Figure~\ref{fig:crossdata}. It is observed that our DICE framework consistently achieves SOTA performance across all four evaluation dimensions compared to other baselines, which indicates that DICE not only enables SLM to effectively acquire domain-specific knowledge but also maintains strong performance across diverse datasets. Notably, when applying models trained on GSM8K to the more challenging MATH test set, DICE demonstrates substantial improvement in C-Acc, exceeding all baselines by at least $35\%$. This significant performance gap underscores DICE's capability to effectively leverage LLM outputs for generating more accurate responses, even when confronted with test data that exhibits greater complexity than the training samples. These findings collectively establish that the DICE framework attains exceptional cross-dataset generalizability relative to existing baseline methods.
\begin{figure*}[ht]
    \centering
    \includegraphics[width=\linewidth]{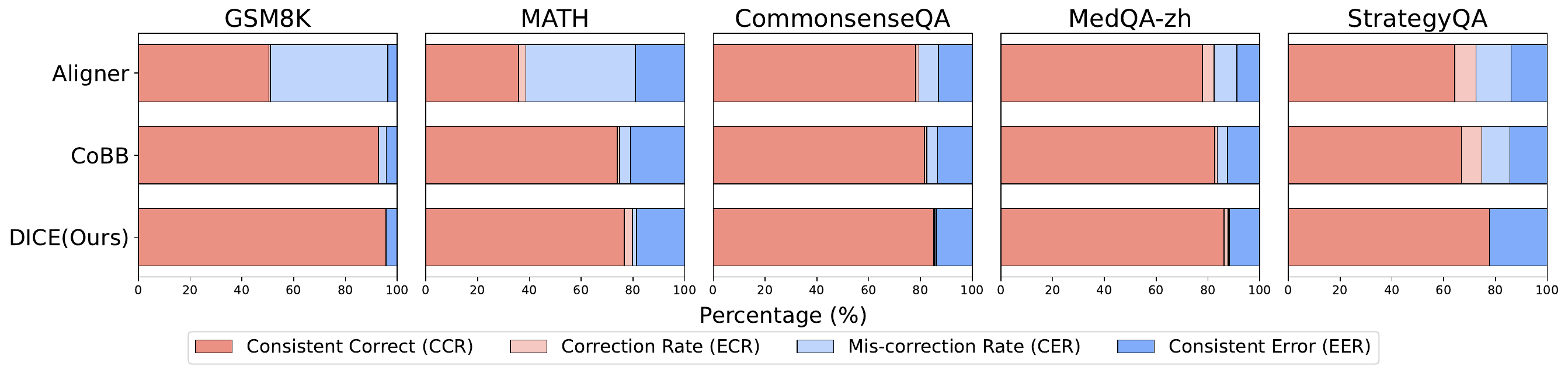}
    \caption{\textbf{The consistency analysis between natural language outputs from LLM and outputs in XML format from 1.5B SLMs in generative approaches.} We investigate the consistency in output correctness using four evaluation metrics: Mis-correction Rate (CER), Correction Rate (ECR), Consistent Error Rate (EER), and Consistent Correct Rate (CCR). These metrics provide a comprehensive insight into the strengths of the DICE framework.}
    \label{fig:consistency}
    \vspace{-1em}
\end{figure*}
 \subsection{Effectiveness Analysis}\label{sec:eff}
To further investigate the insights underlying the effectiveness of our proposed DICE framework, we conduct a comprehensive comparative analysis of consistency between natural language LLM outputs ($y_o$) and the structured outputs from SLM ($\hat{t}$) across different generative model collaboration strategies (Aligner, CoBB, and DICE). We propose four metrics to evaluate the consistency: (1) \textbf{Consistent Correct Rate (CCR)}: The proportion of samples where both $y_o$ and $\hat{t}$ are correct. (2) \textbf{Correction Rate (ECR)}: The proportion of samples where $y_o$ is incorrect but is corrected by the SLM. (3) \textbf{Mis-correction Rate (CER)}: The proportion of samples where $y_o$ is correct but becomes incorrect after adaptation. (4) \textbf{Consistent Error Rate (EER)}: The proportion of samples where both $y_o$ and $\hat{t}$ are incorrect.
The consistency performance of the generative methods is presented in Figure~\ref{fig:consistency}.

The results demonstrate that DICE substantially outperforms both Aligner and CoBB by achieving significantly higher CCR and lower CER. Notably, DICE maintains a CER below $2\%$ across all datasets, indicating its strong ability to preserve the correctness of LLM outputs. This suggests that the analyze-then-answer paradigm adopted in DICE enables the SLM to more effectively utilize the information embedded in LLM outputs without introducing unnecessary modifications. Moreover, for questions beyond LLM’s knowledge coverage, DICE still demonstrates the capability to partially correct erroneous outputs. In contrast, the Aligner and CoBB suffer from ``over-correction dilemma'': though they can achieve relatively high ECR, revealing the ability to correct errors for questions outside the LLM's knowledge domain, they also exhibit high CER which indicates that they fail to adequately analyze and assess the validity of LLM outputs, leading to erroneous modifications and an overall decline in performance.

\subsection{Latency Analysis}
In the inference stage, while our DICE framework and other model collaboration baselines enhance performance, they also introduce latency when compared to LLMs that directly employ ICL. To quantitatively assess this overhead, we compare the inference times of LLM with ICL against various model collaboration approaches. The evaluation experiment is conducted over the same 100 samples from the MATH test set on two NVIDIA A100 GPUs. The model collaboration methods (Aligner, Bbox-Adapter, CoBB, and DICE) use the Qwen2.5-72B-Instruct-GPTQ-Int4 LLM with the fine-tuned Qwen2.5-1.5B-Instruct SLM. To mitigate random variance, each method is evaluated across five independent inference runs, and the average latency is reported as the final result in Table \ref{tab:latency}.
\begin{table}[t]
\renewcommand\arraystretch{1}
\resizebox{1\linewidth}{!}{
\begin{tabular}{lcc}
\toprule
\textbf{Method}    & \textbf{Time}(s/sample)     &\textbf{C-Acc}(\%)     \\\cmidrule{1-3}
LLM (ICL)          & \textbf{0.9191}             & 4.2           \\
LLM + Aligner      & 1.1090                      & 38.6          \\
LLM + BBox-Adapter & 2.2592                      & 9.6           \\
LLM + CoBB         & 1.0854                      & 74.8   \\
LLM + DICE(Ours)   & 1.0999                      & \textbf{79.8} \\
\bottomrule
\end{tabular}
}
\caption{\textbf{Inference time and performance of different methods.} The reported time represents the average inference latency per sample (in seconds) across different methods.}
\label{tab:latency}
\vspace{-1em}
\end{table}

The experimental results indicate that our DICE framework introduces an additional latency of approximately 20\% compared to the LLM with ICL. However, this modest time overhead yields substantial performance gains: content accuracy improves by over 70\% on the MATH test set and by an average of more than 25\% across five different reasoning tasks selected in this work. The significant performance gains achieved with minimal time overhead make this trade-off entirely worthwhile. Furthermore, compared to other model collaboration methods, our DICE framework achieves optimal performance for a comparable computational cost, which demonstrates the high efficiency of our approach.

\subsection{Ablation Study}
\paragraph{Fine-tuning Strategy Ablation}To validate the efficacy of our DICE framework, we conduct the ablation studies by replacing the finetuning procedure with either SFT or the GRPO algorithm exclusively. We train the 1.5B small models for 3 epochs on the structured chain-of-thought adaptation MATH and CSQA benchmarks (more training details can be found in Appendix \ref{ap:impl}). The experimental results are presented in Table \ref{tab:ablation}. It is observed that with the same fine-tuning epochs, the GRPO-only strategy exhibits suboptimal performance in both format adherence and content accuracy, which suggests that the GRPO algorithm is inefficient for models to learn format information from scratch. In contrast, the SFT-only strategy enables the model to effectively learn the user-specified output format and achieve competitive answer C-Acc compared to other baselines in Table \ref{tab:main}, which reveals the effectiveness of the analyze-then-answer generation pattern of data construction in Section~\ref{sec:data-cons}. Moreover, as discussed in Section~\ref{sec:fine-tune}, our dual-tuning strategy that applies GRPO after SFT enables the model to better attend to output format and final answer, achieving the highest F-Acc and C-Acc scores.
\begin{table}[t]
\centering
\renewcommand\arraystretch{1.0}
\resizebox{0.95\linewidth}{!}{
\begin{tabular}{ccccc}
\toprule
\multirow{2}{*}{\textbf{Strategy}} & \multicolumn{2}{c}{\textbf{MATH}} & \multicolumn{2}{c}{\textbf{CSQA}}\\ 
\cmidrule(rl){2-3} \cmidrule(rl){4-5}
    & F-Acc & C-Acc & F-Acc & C-Acc \\ \cmidrule{1-5}
    SFT & 98.4 & 75.6 & \textbf{100.0} & 82.8 \\
    GRPO & 92.0 & 73.2 & 89.2 & 77.2 \\
    SFT+GRPO & \textbf{99.8} & \textbf{78.8} & \textbf{100.0} & \textbf{85.6} \\

\bottomrule
\end{tabular}
}
\caption{\textbf{Ablation experiments on fine-tuning strategy used in DICE framework.} For all experiments, the 1.5B SLMs are fine-tuned for three epochs.}
\label{tab:ablation}
\vspace{-1em}
\end{table}
\paragraph{Correctness Ratio of $y_o$ Ablation} 
% As illustrated in Section \ref{sec:eff}, while the DICE framework significantly reduces the mis-correction rate (CER) compared to baselines, it still struggles with questions that fall outside the LLM’s knowledge scope. This limitation stems from the high proportion of correct LLM outputs in the training data, leading the SLM to overtrust the LLM’s responses, which compromises cross-model generalizability and leads to low correction rate (ECR).
To quantify the impact of the correctness ratio of $y_o$ in the training data and to determine the optimal ratio, we conduct a controlled experiment on the MATH dataset. In this experiment, we maintain a constant training set size of 5,000 samples while varying the proportion of correct $y_o$ across four configurations: 100\%, 75\%, 50\%, and 25\%. All $y_o$ outputs are generated by the Qwen2.5-72B-Instruct-GPTQ-Int4 to ensure consistency. Subsequently, we train four distinct 1.5B SLMs using these training sets and evaluate their performance on the test set by pairing them with various LLMs. The results of this experiment are presented in Appendix \ref{ap:yo}. The experimental results indicate that the higher the proportion of correct $y_o$ in the training set, the more the model tends to inherit the answers from the LLM, thereby reducing miscorrections and enabling better collaboration with stronger LLMs. In contrast, SLMs trained on datasets with lower $y_o$ correctness ratios (e.g., 50\%) exhibit stronger correction capabilities, making them more effective when paired with weaker LLMs. However, this trend does not extend linearly to extreme cases. For instance, training with only 25\% correct $y_o$ resulted in suboptimal performance across all LLM backbones. This is likely because the SLM's inherent capacity limitations prevent it from achieving high correction accuracy when exposed to predominantly incorrect examples. Overall, the model trained with 50\% correct $y_o$ not only achieved the best performance but also exhibited greater stability and robustness.

\section{Conclusion}
In this paper, we propose the DICE framework, a highly efficient and plug-and-play approach that adapts LLMs to structured reasoning tasks. We construct the structured chain-of-thought adaptation datasets that guide SLM to reason before generating final answers. We also design a dual-tuning strategy that leverages the strengths of both SFT and GRPO algorithms. Experimental results demonstrate that DICE achieves near-perfect format adherence while maintaining superior content accuracy, coupled with exceptional generalization capabilities. It addresses the issue of LLMs' instruction-following limitations without directly fine-tuning, achieving a dual enhancement in both reasoning and instruction-following. This innovation holds significant practical value in real-world applications with specific user requirements.

\section*{Limitations}
Although our proposed DICE framework can effectively balance the trade-off between LLMs' instruction-following and reasoning capabilities, demonstrating superior performance in both structure adherence and content accuracy, it nonetheless possesses some limitations. For example, our approach introduces additional computational overhead. For each query, it first invokes LLM to generate natural language outputs, which are then refined by the SLM. However, for relatively simple questions that can be adequately addressed by SLM, invoking LLM is unnecessary and increases both computational cost and latency. Future work could incorporate a mechanism to assess question complexity beforehand, selectively engaging the LLM only when necessary, thereby optimizing resource usage.

\section*{Ethical Considerations}
All models utilized in this study, except for GPT-4.1-mini, and all datasets are open-source. We downloaded the open-source models from their official releases on Hugging Face and accessed GPT models via the OpenAI API. Throughout, we strictly comply with all applicable user licenses. The datasets utilized in this research are sourced from the officially published repositories and are used exclusively for academic research purposes. All datasets are widely used and contain no personal or sensitive information. Therefore, there is no risk of personal information leakage here.

For AI usage, we only use AI assistants to check typos and grammar errors when writing.

\section*{Acknowledgements}
This work was supported by the National Key R\&D Program of China (No. 2022ZD0162101).
\bibliography{custom}

\appendix
\section{Datasets and Metrics}\label{sec:data&metric}
This section presents further details about the datasets and the two evaluation metrics used in the work.
\subsection{Datasets}\label{sec:data}
In the experiments of this work,  we select five commonly used datasets from four types of reasoning tasks: mathematical reasoning, commonsense reasoning, task-specific reasoning, and implicit reasoning.
\begin{itemize}
    \item \textbf{GSM8K} \cite{cobbe2021gsm8k}, short for ``Grade School Math 8K'', is a benchmark consisting of 8,500 question-answering math problems designed to evaluate the fundamental mathematical reasoning abilities of models. In this dataset, each problem is paired with step-by-step reasoning and the numerical answer.
    \item \textbf{MATH} \cite{hendrycks2021math} is a more challenging question-answering dataset consisting of 12,500 algebra, calculus, geometry, and precalculus problems sourced from high school mathematics competitions. Compared to GSM8K, MATH requires deeper mathematical knowledge and more sophisticated multi-step problem-solving capabilities.
    \item \textbf{CommonsenseQA} \cite{talmor2018commonsenseqa} is a multiple-choice question-answering benchmark designed to evaluate model's commonsense reasoning ability. It contains 12.1k questions sourced from ConceptNet, requiring models to leverage general world knowledge and contextual understanding to select the correct answers from the given five options.
    \item \textbf{MedQA} \cite{jin2021medqa} is a domain-specific dataset containing 61,097 multiple-choice questions from the United States Medical Licensing Examination (English), Chinese National Medical Licensing Exam (simplified Chinese), and Taiwan’s Medical Licensing Exam (traditional Chinese). In this paper, we only sample data from the simplified Chinese subset to enrich the language diversity of our experiment, so we denote the dataset as MedQA-zh.
    \item \textbf{StrategyQA} \cite{geva2021strategyqa} is a binary benchmark for evaluating model's strategic reasoning and implicit task understanding capabilities. It includes 2,290 True/False questions that require models to decompose complex problems into executable reasoning steps. 
\end{itemize}

Notably, to reduce computational costs, we randomly sample 1,000 training examples and 500 test examples from each dataset as our original training and evaluation sets (the test set of StrategyQA contains only 229 examples). When constructing the new structured chain-of-thought dataset illustrated in Section~\ref{sec:data-cons}, we prompt the LLM to sample five responses for each training sample. This allows us to expand the size of the training set to approximately 5,000 examples without increasing the number of LLM invocations, although some samples are filtered out after the two-stage rationale generation process.
\subsection{Metrics}\label{sec:metrics}
We design two metrics: format accuracy and content accuracy, to respectively evaluate the model's format adherence and reasoning capability:
\begin{itemize}
    \item \textbf{Format accuracy (F-Acc)}: Proportion of samples with correct output format in all outputs. We employ string matching to extract the model outputs. An output is classified as having correct formatting only if it contains all specified keywords in the required format or can be automatically converted into the specific structure using standard toolkits.
    \item \textbf{Content accuracy (C-Acc)}: Content accuracy is measured by calculating the Exact Match (EM) score between the extracted final answer from the model's response and the ground truth. Notably, final answers can only be fully extracted when the output format is correct; for format-incorrect samples, their final answers remain unextractable and are therefore judged as incorrect. Consequently, any output with correct content must necessarily have a correct format.
\end{itemize}

\section{Format Details}\label{sec:format}
\begin{figure}[t]
    \centering
    \includegraphics[width=\linewidth]{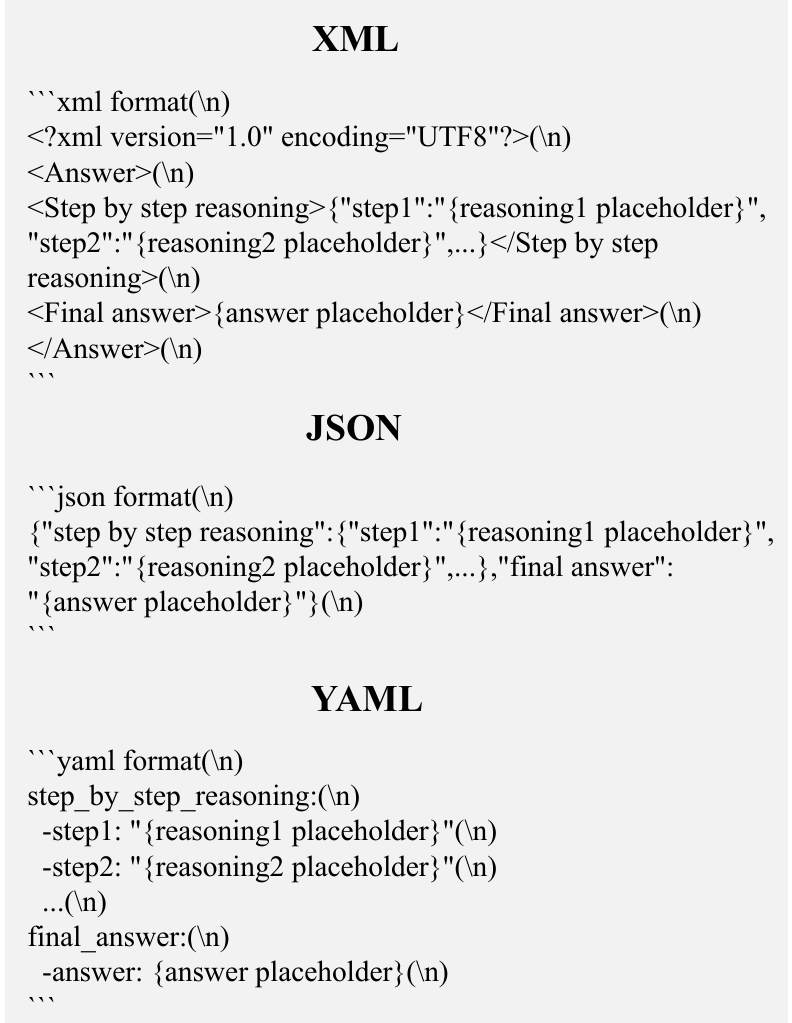}
    \caption{Format templates for GSM8K, MATH, and StrategyQA datasets. The symbol  \texttt{(\textbackslash n)} indicates the presence of a newline character at that specific position.}
    \label{fig:template1}
\end{figure}
\begin{figure}[ht]
    \centering
    \includegraphics[width=\linewidth]{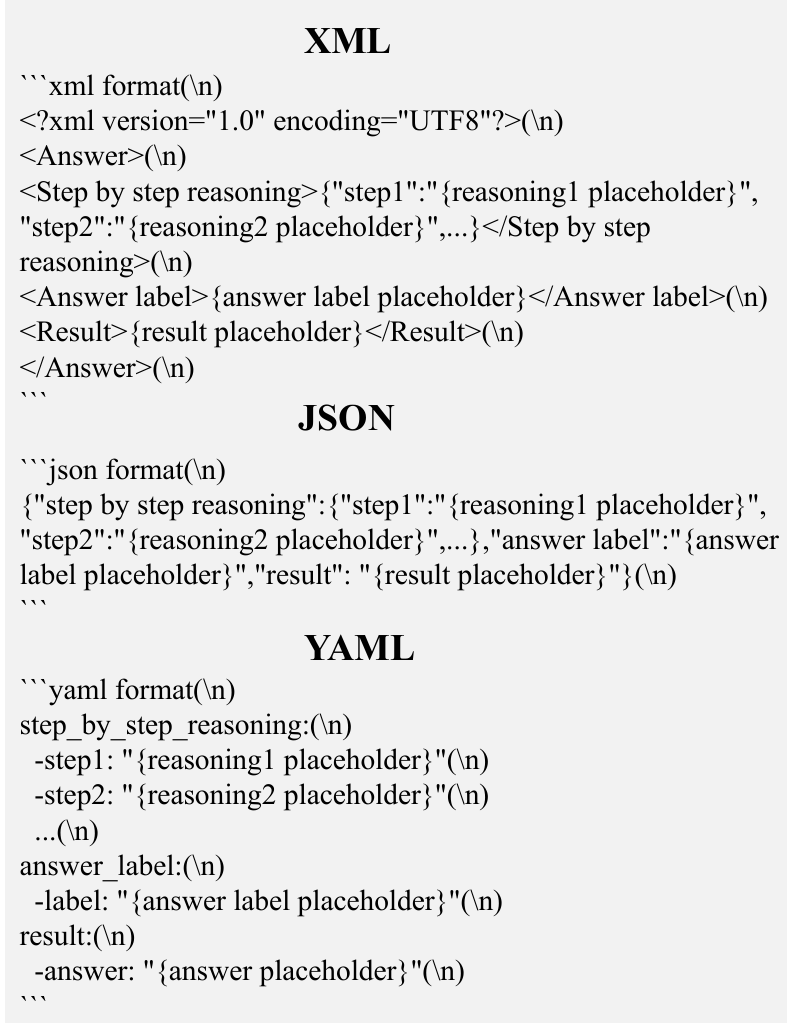}
    \caption{Format templates for CommonsenseQA and MedQA datasets. The symbol  \texttt{(\textbackslash n)} indicates the presence of a newline character at that specific position.}
    \label{fig:template2}
\end{figure}
\begin{figure*}[ht]
    \centering
    \includegraphics[width=\linewidth]{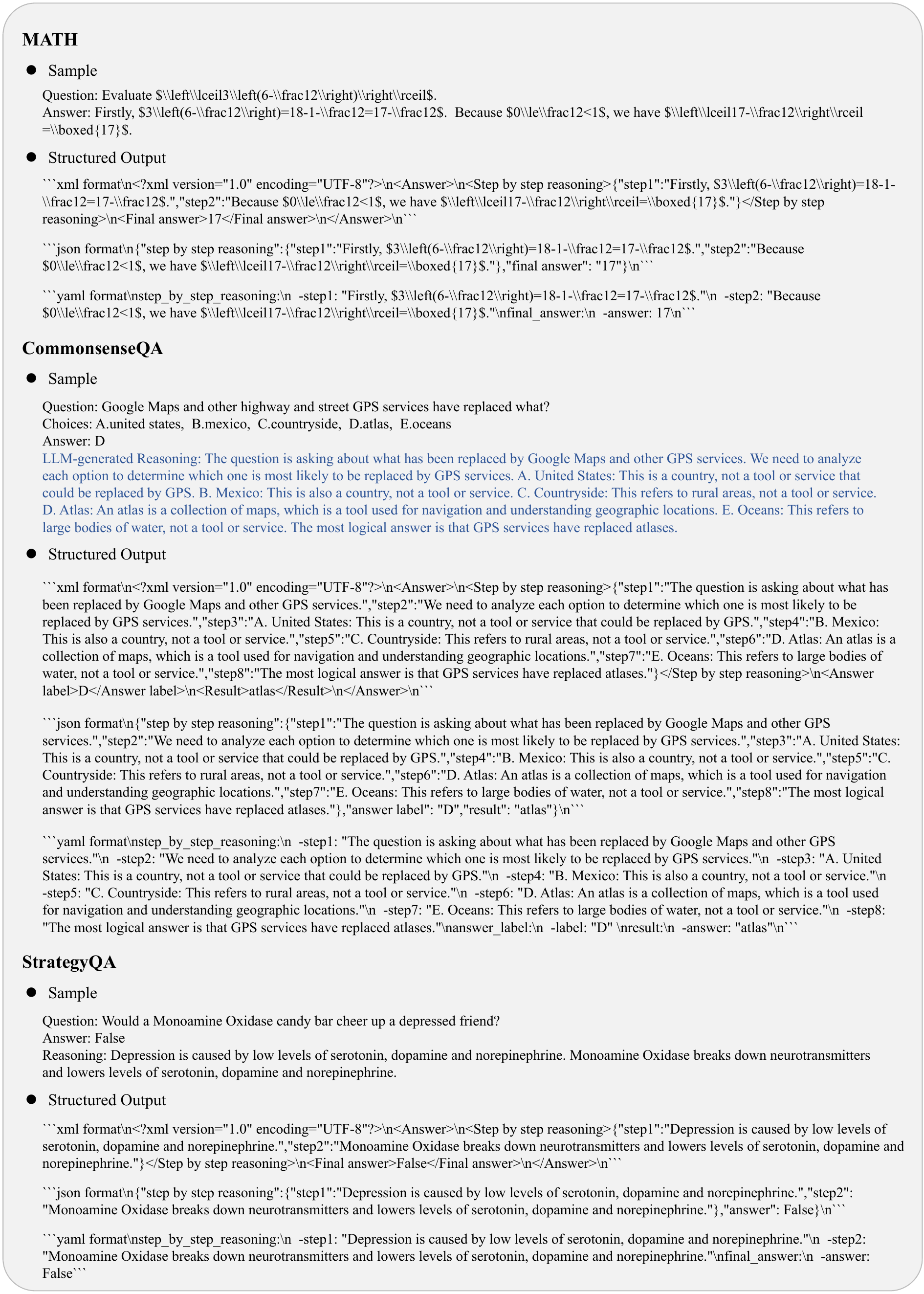}
    \caption{Examples from MATH, CommonsenseQA, and StrategyQA. The original CommonsenseQA dataset does not contain reasoning information; therefore, we instruct the LLM to generate reasoning. }
    \label{fig:format}
\end{figure*}

In our experiment, we employed three formats: XML, JSON, and YAML. 
For the three question-answering datasets, including GSM8K, MATH, and StrategyQA, the formatted output incorporates step-by-step reasoning along with the final answer. For multiple-choice datasets including CommonsenseQA and MedQA, the model is additionally required to generate the selected option label along with its corresponding answer, in addition to the reasoning steps. The basic templates of the three formats for GSM8K, MATH, StrategyQA and CommonsenseQA, MedQA are shown in Figure~\ref{fig:template1} and \ref{fig:template2}, respectively. As illustrated in Figure~\ref{fig:format}, we provide representative examples from MATH, CommonsenseQA, and StrategyQA. 

\section{Algorithm Details of DICE}\label{ap:algorithm}
In this section, we elaborate on the core details of the GRPO algorithm and present the complete DICE algorithm in Algorithm \ref{alg:AOS}.

\subsection{GRPO Algorithm Details}\label{ap:grpo}
In the fine-tuning process of LLMs, reinforcement learning (RL) plays a pivotal role \cite{schulman2017ppo,rafailov2024dpo,azar2023IPO,ethayarajh2024kto}. Although the traditional Proximal Policy Optimization (PPO \cite{schulman2017ppo}) algorithm has been widely adopted for LLM fine-tuning, it requires maintaining a separate value network comparable in size to the policy model for advantage function estimation, leading to substantial memory consumption and computational overhead in large-scale scenarios. To address these challenges, the Group Relative Policy Optimization (GRPO \cite{shao2024grpo}) algorithm is proposed, which seeks to minimize dependence on value networks while preserving the stability and efficiency of policy updates.

The GRPO framework operates by sampling a group of actions from the current policy and calculating relative advantages within this group, thereby eliminating the need for a critic model. The advantage estimation can be formulated as:
\begin{equation}
    A_i=\frac{r_i-\mathrm{mean}(\{r_1,r_2,\cdots,r_G\})}{\mathrm{std}(\{r_1,r_2,\cdots,r_G\})}
\end{equation}
where G denotes the group size (number of sampled actions per iteration). The complete loss function of GRPO can be expressed as:
\begin{equation}\label{eq:grpo}
\resizebox{.88\hsize}{!}{$
    \begin{aligned}
        \mathcal{L}_{GRPO}(\theta) & =\frac{1}{G} \sum_{i=1}^{G}\bigg(\min\bigg(\frac{\pi_{\theta}\left(o_{i}\right)}{\pi_{\theta_{\mathrm{old}}}\left(o_{i}\right)}A_{i}, \mathrm{clip}\bigg(\frac{\pi_{\theta}\left(o_{i}\right)}{\pi_{\theta_{\mathrm{old}}}\left(o_{i}\right)},\\
        & 1-\varepsilon, 1+\varepsilon\bigg)A_{i}\bigg)  - \beta \mathbb{D}_{KL}\left(\pi_{\theta}\parallel\pi_{\mathrm{ref}}\right)\bigg)
\end{aligned}
 $}
\end{equation}
\subsection{Overview of DICE Algorithm}
The overview algorithm of our proposed DICE framework is shown in Algorithm \ref{alg:AOS}.
\SetCommentSty{mycommfont}
\newcommand{\mycommfont}[1]{\textcolor{gray}{\texttt{#1}}}

\section{Implementation Details}\label{sec:implementation}
\subsection{Structured Data Construction}
For SFT and Aligner, we constructed the training sets using only the original data. Since the CommonsenseQA and MedQA datasets provide only the answers without any reasoning information, the target structured outputs for these two datasets in SFT and Aligner contain only the selected option label and the corresponding answer, without step-by-step reasoning. The step-by-step reasoning in GSM8K, MATH, and StrategyQA is derived from the original benchmarks.

For BBox-Adapter, we first prompt the Qwen2.5-72B-Instruct-GPTQ-Int4 to generate reasoning for each training sample in CommonsenseQA and MedQA. Then we use the reasoning and ground-truth answer of all five benchmarks to construct standard structured outputs. Subsequently, we instruct the LLM to generate five candidates for each question in the training set with a 2-shot prompt. The small model was then trained using both the standard structured outputs and candidates.

For CoBB, we generate positive reasoning via Qwen2.5-72B-Instruct-GPTQ-Int4 and randomly sample reasoning from other questions as negative reasoning. This approach allowed us to construct both positive and negative structured outputs.
\subsection{Fine-tuning Details}\label{ap:impl}
Our experiments were conducted on NVIDIA A100 GPUs (80G memory). When running all baselines and our proposed DICE method, we utilized the vLLM \cite{kwon2023vllm}, LLama-Factory \cite{zheng2024llamafactory}, and SWIFT \cite{zhao2024swift} frameworks for model fine-tuning.

\paragraph{Main Experiment} In the main experiments (Tables \ref{tab:main} and \ref{tab:dif-formats}), we use Qwen2.5-72B-Instruct-GPTQ-Int4 as the large language model, and Qwen2.5-0.5B-Instruct, Qwen2.5-1.5B-Instruct, and Qwen2.5-3B-Instruct as the small models. All models are trained on two A100 GPUs with bf16 precision. Other detailed experimental configurations are: (1) For SFT and Aligner baselines, LoRA fine-tuning is applied with rank 32 and alpha 64 for 3 epochs. The training process uses a batch size of 64, learning rate of $2\times10^{-4}$, warmup ratio of 0.1, and weight decay of 0.1. (2) BBox-Adapter adopts full fine-tuning for 3 epochs. During training, the candidate count is set to 5, max length to 1, batch size to 100, and the model operates in classification mode. The remaining parameters are kept consistent with the original paper. (3) CoBB utilizes LoRA rank 32 and alpha 64 for 5 epochs. The hyperparameter $\lambda$ is fixed at 0.1, while the training employs a batch size of 64 and a learning rate of $1\times10^{-5}$, maintaining other parameters as in the original work. (4) For our proposed DICE framework, both SFT and GRPO stages adopt LoRA rank 32 and alpha 64. During the SFT stage, the SLM is trained for 2 epochs with batch size 64, learning rate $2\times10^{-4}$, warmup ratio 0.1, and weight decay 0.1. In the subsequent GRPO stage, the hyperparameter $G$ is set to 16, the learning rate is reduced to $1\times10^{-5}$, temperature is set to 0.8, batch size increases to 128, warmup ratio decreases to 0.05, and training for 1 additional epoch.
\paragraph{Ablation Study} In the ablation study, we utilize Qwen2.5-72B-Instruct-GPTQ-Int4 as the LLM and Qwen2.5-1.5B-Instruct as the original SLM. All experiments employ LoRA with a rank of 32 and an alpha of 64, conducted at bf16 precision. In the SFT-only setting, we set the learning rate to 2e-4, batch size to 64, warmup ratio and weight decay to 0.1, and train for 3 epochs. In the GRPO-only setting, we set the hyperparameter $G$ to 16, learning rate to 2e-5, batch size to 128, temperature to 0.9, and also train for 3 epochs. All other hyperparameters remain consistent with those used in the main experiments.

\section{Correctness Ratio of $y_o$ Ablation Experiment Result}\label{ap:yo}
The content accuracy of models trained with different correctness ratio of $y_o$ and inferenced with different LLMs are illustrated in Table \ref{tab:yo}.
\begin{table}[ht]
\renewcommand\arraystretch{1}
\resizebox{1\linewidth}{!}{
\begin{tabular}{lcccc}
\toprule
\textbf{Ratio}        & \textbf{100\%} & \textbf{75\%}  & \textbf{50\%}  & \textbf{25\%} \\ \cmidrule{1-5}
Qwen2.5-72B + DICE  & 79.0  & \textbf{79.4} & 78.0          & 74.6 \\
Qwen2.5-7B + DICE   & 71.8  & 70.6          & \textbf{72.0} & 67.8 \\
Llama3-8B + DICE    & 33.6  & 33.0          & \textbf{34.4} & 34.0 \\
GPT-4.1-mini + DICE & 67.8  & 68.8          & \textbf{70.8} & 65.4 \\
Average      & 63.1  & 63.0          & \textbf{63.8} & 60.5 \\
\bottomrule
\end{tabular}
}
\caption{\textbf{Ablation experiments on the correctness ratio of $y_o$.} The models' name Qwen2.5-72B, Qwen2.5-7B, and Llama3-8B are short for Qwen2.5-72B-Instruct-GPTQ-Int4, Qwen2.5-7B-Instruct, and Meta-Llama-3-8B-Instruct.}
\label{tab:yo}
\end{table}
\end{document}